\documentclass{article} 
\usepackage{iclr2026_conference,times}

\usepackage{amsmath,amsfonts,bm}

\def\eqref#1{equation~\ref{#1}}

\def\1{\bm{1}}

\DeclareMathAlphabet{\mathsfit}{\encodingdefault}{\sfdefault}{m}{sl}
\SetMathAlphabet{\mathsfit}{bold}{\encodingdefault}{\sfdefault}{bx}{n}

\usepackage{hyperref}
\usepackage{url}

\usepackage{algorithm} 
\usepackage{algorithmic}

\usepackage{graphicx}
\usepackage[utf8]{inputenc}
\usepackage{booktabs}
\usepackage{multirow}
\usepackage[table]{xcolor}
\usepackage{amsmath}
\usepackage{enumitem}
\usepackage{tabularx}

\newcolumntype{Y}{>{\centering\arraybackslash}X}

\definecolor{header_color}{RGB}{235, 235, 235}
\definecolor{best_cell}{RGB}{220, 240, 220}

\title{PhysNote: Self-Knowledge Notes for Evolvable Physical Reasoning in Vision-Language Model}
\iclrfinalcopy

\author{\vspace{-0.1in}
    \textbf{Sinin Zhang}$^{1}$, 
    \textbf{Yunfei Xie}$^{2}$, 
    \textbf{Yuxuan Cheng}$^{1\dagger}$%
    \makeatletter{\footnotetext{\hspace{-7pt}$^{\dagger}$Corresponding author: yuxuancheng@link.cuhk.edu.cn.}\makeatother}, 
    \textbf{Haoyu Zhang}$^{3}$,
    \textbf{Tong Zhang}$^{4}$\vspace{0.1in}\\
    $^1$The Chinese University of Hong Kong, Shenzhen, 
    $^2$Rice University, \\ $^3$City University of Hong Kong, $^4$Fudan University\vspace{0.05in}\\
}
\begin{document}

\maketitle

\begin{abstract}
Vision-Language Models (VLMs) have demonstrated strong performance on textbook-style physics problems, yet they frequently fail when confronted with dynamic real-world scenarios that require temporal consistency and causal reasoning across frames. We identify two fundamental challenges underlying these failures: (1) spatio-temporal identity drift, where objects lose their physical identity across successive frames and break causal chains, and (2) volatility of inference-time insights, where a model may occasionally produce correct physical reasoning but never consolidates it for future reuse. To address these challenges, we propose PhysNote, an agentic framework that enables VLMs to externalize and refine physical knowledge through self-generated ``Knowledge Notes.'' PhysNote stabilizes dynamic perception through spatio-temporal canonicalization, organizes self-generated insights into a hierarchical knowledge repository, and drives an iterative reasoning loop that grounds hypotheses in visual evidence before consolidating verified knowledge. Experiments on PhysBench demonstrate that PhysNote achieves 56.68\% overall accuracy, a 4.96\% improvement over the best multi-agent baseline, with consistent gains across all four physical reasoning domains. 

\end{abstract}

\section{Introduction}

Building autonomous agents that can act safely in the physical world requires \emph{physical reasoning}: the ability to connect observations with the causal mechanisms that generate motion, contact, and state changes over time. Humans acquire this ability early as intuitive physics and apply it effortlessly, yet replicating it in artificial systems remains an open challenge. This capability links visual perception to prediction and control, and it is therefore central to reliable embodied interaction, from robotic manipulation to autonomous navigation.

Recent Vision-Language Models (VLMs) have shown strong performance on static scene understanding and textbook-style physics problems~\cite{he2024olympiadbench, yue2024mmmu}. With increased pre-training scale, models~\cite{achiam2023gpt, guo2025deepseek, comanici2025gemini} can often infer mathematical solutions from single-frame schematics and recognize fundamental physical concepts. However, much of this progress reflects competence in ``text-based physics''~\cite{qiu2025phybench, zhang2025physreason}, where the task is presented as a symbolic or linguistic problem and the supervision signal is contained in the prompt. In such settings, correct answers can be produced without constructing a temporally grounded representation of the underlying physical process~\cite{chow2025physbench, yang2025cambrian}. This limitation becomes clear when the input shifts from static images to multi-frame sequences that contain temporal evolution and causal interaction among objects~\cite{lecun2022path}. In these dynamic settings, the model must track objects across time, detect contact events, and maintain consistent causal chains, capabilities that static, text-based tasks never require.

Even state-of-the-art VLMs often fail at these requirements. An extensive evaluation of 75 VLMs on PhysBench~\cite{chow2025physbench} shows that most models achieve only around 40\% accuracy on physical reasoning tasks, far below human performance, with perceptual errors and knowledge gaps accounting for the majority of failures. Notably, this deficit does not diminish with larger model size, more training data, or additional input frames, suggesting that physical understanding is not an emergent property of scale alone. Objects frequently lose their physical identity across successive frames, and causal chains can be broken by hallucinatory transitions. These findings indicate that current VLMs lack mechanisms to reliably anchor perception and reasoning in both time and space, which limits their applicability to embodied tasks that demand precise physical inference, such as manipulation where misidentifying an object or misjudging a trajectory can cause operational failures~\cite{chow2025physbench}.

We identify two fundamental challenges that underlie these failures in dynamic physical reasoning: (1) \textbf{Spatio-Temporal Identity Drift in Perception} and (2) \textbf{Volatility of Inference-time Insights}.

For the first challenge, dynamic physical reasoning requires object permanence and traceability of states across time. However, standard VLMs frequently exhibit ``identity drift,'' in which the semantic representation of an object shifts or disappears across successive frames. Without an explicit mechanism for spatio-temporal anchoring, a causal chain, such as a collision that leads to a particular trajectory, becomes fragmented, which prevents the model from forming a coherent representation of scene dynamics.

For the second challenge, human physicists do not treat every new problem as a blank slate; they externalize insights, refine heuristics, and consolidate knowledge into reusable notes. In contrast, existing reasoning paradigms in VLMs are largely \emph{volatile}. While a model may occasionally generate profound physical insights during a specific reasoning trace, these insights are transient and are not consolidated into a persistent knowledge store. Consequently, the model remains a ``goldfish-memory physicist,'' failing to evolve its internal world model through experience or self-reflection.

To address these challenges, we introduce \textbf{PhysNote}, an agentic framework that enables VLMs to externalize and refine physical knowledge through self-generated \textbf{``Knowledge Notes.''} PhysNote is designed to support dynamic reasoning by iterating over observation of a sequence, retrieval of relevant prior notes, and synthesis of updated insights that remain available for future problems.

PhysNote comprises three core components. To mitigate Spatio-Temporal Identity Drift in Perception, PhysNote employs a \textbf{Spatio-Temporal Canonicalization} module (Sec.~\ref{sec:perception}) that assigns immutable numeric identifiers to visual tokens and enforces a \emph{Triadic Observation Template}, providing stable visual anchors for object permanence across temporal sequences. To mitigate Volatility of Inference-time Insights, PhysNote introduces a \textbf{Hierarchical Knowledge Architecture} (Sec.~\ref{sec:hierarchy}) that externalizes self-generated insights into a three-tier repository spanning domain-level strategies and task-specific procedural details. This repository is updated through autonomous pruning and reflection, which preserves high-fidelity knowledge while systematically discarding erroneous biases. Connecting these two components, an \textbf{Agentic Reasoning} engine (Sec.~\ref{sec:infoagent}) drives an iterative Hypothesis-Evidence-Validation loop: the agent gathers grounded visual evidence through the canonicalized anchors, consults the Knowledge Notes for relevant physical heuristics, and consolidates verified insights back into the repository after each problem.

Experiments on PhysBench~\cite{chow2025physbench} show consistent gains for open-source models. On the large-scale test set, PhysNote achieves 56.68\% overall accuracy, a 4.96\% absolute improvement over the best multi-agent baseline, with balanced gains across all four physical domains. On the validation set, PhysNote reaches 72.86\%, exceeding the Qwen2.5-VL-72B baseline by 3.01\%. Ablation studies confirm that the synergy between the InfoAgent and Knowledge Notes is essential: neither component alone fully accounts for the gain. Qualitative analysis further shows that PhysNote prevents identity drift in dynamic scenes and enables the model to detect fine-grained physical cues that standard VLMs overlook.

\section{Related Work}

\subsection{Physical Reasoning Models}
The quest for physical intelligence in AI has evolved from symbolic engines to data-driven neural architectures. Early works primarily focused on Physics-Specialized Models~\cite{guen2020disentangling, duan2022pip}, which utilize neural networks to approximate differential equations or Lagrangian dynamics. While precise, these models are often task-specific (e.g., fluid dynamics or rigid-body collisions) and require structured representations that are unavailable in raw visual inputs.

Recently, the focus has shifted to Vision-Language Models (VLMs) as generalist observers~\cite{xu2024penetrative}. However, these models often suffer from Spatio-Temporal Identity Drift~\cite{yang2025cambrian}. VLMs lack spatial reasoning and temporal grounding, causing objects to lose their physical identity across sequences. In contrast, PhysNote addresses this by replacing implicit visual attention with a deterministic Spatio-Temporal Canonicalization module, ensuring that visual anchors remain consistent throughout the reasoning trajectory.

\subsection{Agent-Based Physical Understanding}
The integration of Vision-Language Models (VLMs) into Autonomous Agents has enabled models to tackle complex physical tasks via iterative prompting, such as Chain-of-Thought~\cite{wei2022chain} or ReAct paradigms \cite{yao2022react}. To enhance physical grounding, specialized frameworks like PhysAgent \cite{chow2025physbench} incorporate a ``reason-act-observe'' loop that interfaces with external visual tools (e.g., SAM~\cite{kirillov_2023_segment}, Depth Anything~\cite{yang_2024_depth}) to extract precise geometric and state information. Alternatively, Physics Context Builders (PCBs) \cite{balazadeh2025physics} employ modular, fine-tuned smaller VLMs to generate detailed scene descriptions that serve as auxiliary reasoning context.
Despite their successes, these approaches either rely on ephemeral reasoning traces that dissipate once the context window closes, or require static fine-tuning that lacks the capacity for autonomous knowledge evolution.

In contrast, PhysNote enables the persistent externalization and autonomous refinement of physical heuristics across a broader spectrum of dynamic scenarios without the need for expensive parameter fine-tuning or specialized external tools.

\begin{figure*}[t]
    \centering
    \includegraphics[width=\textwidth]{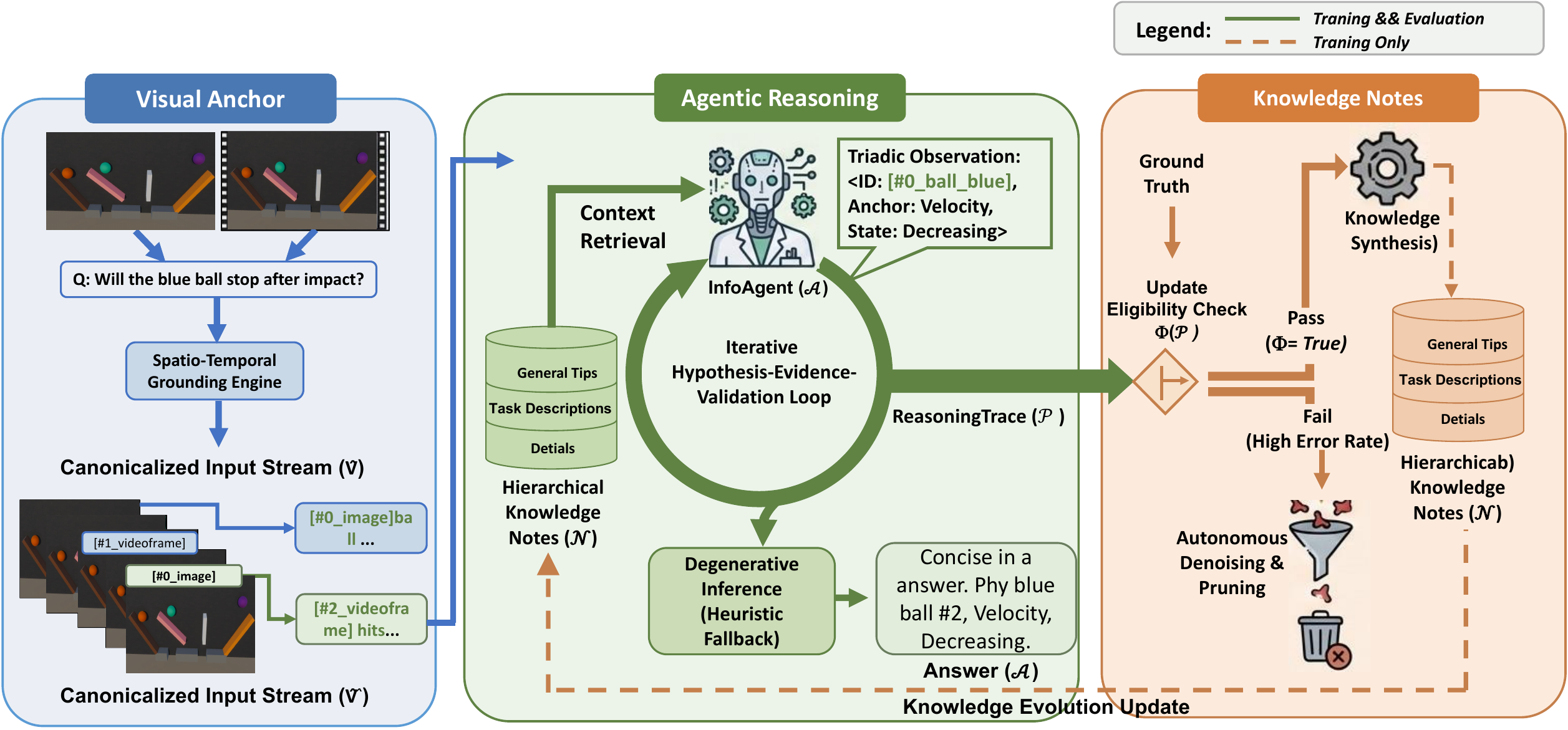}
    \caption{Overview of the PhysNote framework, which operates across three interconnected spaces to enable evolvable physical reasoning. \textbf{Visual Anchors} (left): Given a question $Q$ and visual inputs $V$ (images or video), a Spatio-Temporal Grounding Engine assigns each visual entity an immutable identifier (e.g., \texttt{[\#0\_ball\_blue]}) to produce a canonicalized input stream $\hat{V}$, establishing object permanence across frames. \textbf{Agentic Reasoning} (center): The InfoAgent $\mathcal{A}$ retrieves task-relevant context from Hierarchical Knowledge Notes $\mathcal{N}$, organized into General Tips, Task Descriptions, and Details. It then executes an iterative Hypothesis-Evidence-Validation loop, extracting structured Triadic Observations $\langle \text{ID}, \text{Anchor}, \text{State} \rangle$ from the visual data to build a grounded reasoning trace $\mathcal{P}$. If the evidence gap remains unclosed, a Degenerative Inference mode applies heuristic fallback to produce the final answer $\mathcal{A}$. \textbf{Knowledge Notes} (right, training only): An Update Eligibility Check $\Phi(\mathcal{P})$ evaluates whether the reasoning trace meets the criteria for knowledge consolidation. Eligible traces ($\Phi = \textit{True}$) are synthesized into updated Knowledge Notes; task nodes with persistently high error rates are removed through Autonomous Denoising and Pruning, ensuring the knowledge base evolves toward higher fidelity.}
    \label{fig:framework}
\end{figure*}

\section{Methodology}
\label{sec:methodology}

Current Vision-Language Models (VLMs) can analyze single-frame attributes and solve text-based physics problems, yet they frequently fail in multi-frame dynamic scenarios that demand temporal consistency and causal modeling. This failure stems not from a lack of internal physical knowledge, but from an inability to connect abstract principles to raw, noisy visual signals, much like a skilled mechanic who still requires a \textit{checklist} to inspect the right components in the right order.

To bridge this gap, we propose \textit{PhysNote}, a multi-agent framework that enables a VLM to generate, consult, and refine its own ``Knowledge Notes.'' As illustrated in Figure~\ref{fig:framework}, the framework operates through a structured pipeline: it first organizes raw visual data by assigning unique identifiers, then identifies the task type and retrieves relevant notes, executes an iterative reasoning loop, and finally reflects on the outcome to update the notes. This pipeline comprises three core stages: (1)~\textit{\textbf{Spatio-Temporal Canonicalization}} (Section~\ref{sec:perception}), which creates stable visual anchors; (2)~\textit{\textbf{Hierarchical Knowledge Architecture}} (Section~\ref{sec:hierarchy}), which stores task-specific insights; and (3)~\textit{\textbf{Agentic Reasoning}} (Section~\ref{sec:infoagent}), which performs information discovery and knowledge consolidation.

\begin{figure*}[t]
    \centering
    \includegraphics[width=0.95\textwidth]{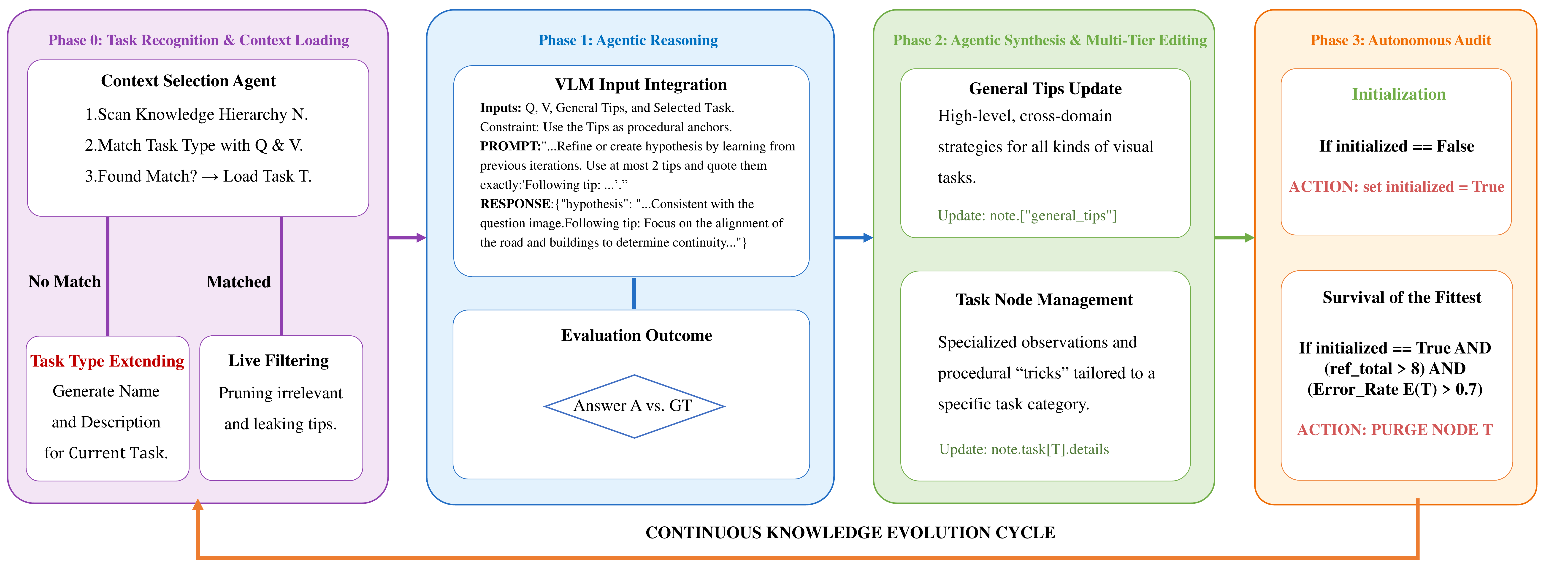}
    \caption{The Knowledge Note pipeline: usage and evolution. During inference, the system retrieves task-relevant notes from the Hierarchical Knowledge Notes $\mathcal{N}$ based on the canonicalized inputs $(\hat{Q}, \hat{V})$, and feeds them into the InfoAgent's iterative Hypothesis-Evidence-Validation loop to produce a grounded answer. During training, the resulting reasoning trace $\mathcal{P}$ undergoes an Update Eligibility Check $\Phi(\mathcal{P})$: eligible traces trigger knowledge synthesis to update the notes, while task nodes with persistently high error rates are pruned, enabling the knowledge base to evolve autonomously toward higher fidelity over successive training batches.}
    \label{fig:note_pipeline}
\end{figure*}

\subsection{Visual Anchors: Spatio-Temporal Canonicalization}
\label{sec:perception}

A fundamental prerequisite for physical reasoning is object permanence: the ability to track an entity across time. Conventional VLMs often struggle with ``identity drift,'' where an object is recognized differently from one frame to the next. To mitigate this, given a raw question $Q$ and its associated visual inputs $V$ (images or video frames), we implement a canonicalization pipeline that transforms $(Q, V)$ into a grounded reference pair $(\hat{Q}, \hat{V})$.

\paragraph{Spatio-Temporal Indexing.} For any input $V$, the system discriminates between static images and video sequences to assign immutable identifiers, such as \texttt{[\#0image]} or \texttt{[\#1videoframe3]}. These identifiers are embedded into the original question text to produce $\hat{Q}$, so that every reference to a visual element is explicit and unambiguous. The visual inputs are likewise repackaged with their assigned identifiers to produce $\hat{V}$. Together, the tagged visuals $\hat{V}$ and the annotated question $\hat{Q}$ establish a consistent referencing scheme for all downstream processing.

\paragraph{Triadic Observation Template.} While Spatio-Temporal Indexing resolves temporal ambiguity by tagging each frame, an additional mechanism is needed to resolve entity ambiguity within each frame. To bridge the gap between raw pixels and symbolic reasoning, we require every observation to answer three questions: \textit{which frame} it comes from, \textit{which object} it describes, and \textit{what property} is observed. Every observation $O$ extracted by the InfoAgent (Section~\ref{sec:infoagent}) from $\hat{V}$ is accordingly formalized as:
\begin{equation}
\label{eq:triplet}
O = \bigl(\, \underbrace{v}_{\text{Visual ID}},\; \underbrace{e}_{\text{Visual Anchor}},\; \underbrace{d}_{\text{Description}} \,\bigr)
\end{equation}
Here $v$ is the identifier assigned during Spatio-Temporal Indexing, $e$ denotes the physical entity of interest (e.g., a specific ball or surface), and $d$ states a single observable property of that entity (e.g., position, velocity, or contact state).

By enforcing this structured format, every observation is traceable to a specific frame and entity, preventing information loss when observations are passed between processing stages (Section~\ref{sec:infoagent}). This protocol compels the model to decompose complex dynamic scenes into a series of verifiable, grounded facts rather than vague or misidentified references.

\subsection{Knowledge Hierarchy and Task-Oriented Selection}
\label{sec:hierarchy}

To store and retrieve physical insights, we externalize self-generated knowledge into a three-tier hierarchy $\mathcal{N}$, termed \textit{Knowledge Notes}. This repository is structured to balance universal physical principles with task-specific heuristics, comprising three levels. (1)~\textbf{General Tips} capture high-level, cross-domain strategies that apply broadly across different physical scenarios. (2)~\textbf{Task Descriptions} provide natural-language definitions that delineate the boundaries of a specific physical phenomenon, specifying both textual and visual cues that characterize the task category. (3)~\textbf{Task Details} record specialized observations and procedural guidance tailored to each task category, accumulated through the reflective process described in Section~\ref{sec:infoagent}.

The selection process is governed by a \textit{Context Selection Agent}. Given the canonicalized inputs $(\hat{Q}, \hat{V})$, the agent evaluates them against $\mathcal{N}$ to identify the most relevant task node $T$. Unlike traditional retrieval, this agent operates with \textit{Global Visibility}, meaning it can synthesize information across the entire hierarchy to determine if the current scenario matches an existing node or represents a novel physical task. If the system fails to find a suitable match and the expansion mode is active, it invokes a \textit{Note Discovery} routine to define a new task node $T$. By observing the global state of $\mathcal{N}$, the system avoids redundant definitions, ensuring that categories such as ``Free Fall'' and ``Projectile Motion'' remain semantically distinct yet functionally related.

To ensure the long-term reliability of $\mathcal{N}$ and prevent ``knowledge pollution'' from erroneous reasoning traces, we implement an autonomous pruning mechanism. The core idea is simple: every time a knowledge node is consulted, the system records whether the inference succeeded or failed. For each task node $T$, let $n_+$ and $n_-$ denote the cumulative counts of correct and incorrect outcomes. The empirical error rate is:
\begin{equation}
E(T) = \frac{n_-}{n_+ + n_-}
\end{equation}
Once $E(T)$ exceeds a predefined threshold $\tau$ after a minimum of $n_{\min}$ activations, the node $T$ and its associated \textit{Task Details} are purged from $\mathcal{N}$. This ``survival of the fittest'' logic ensures that the knowledge base evolves towards higher precision, systematically discarding speculative or biased information that may have been generated during earlier iterations.

\subsection{Agentic Reasoning: The InfoAgent Loop and Reflection}
\label{sec:infoagent}

The core reasoning engine, the \textit{InfoAgent}, is designed to transform high-noise visual environments into structured, textbook-style physics problems. This is achieved through an iterative \textit{Hypothesis--Evidence--Validation} loop, which is capped at a maximum of three iterations ($i \in \{1, 2, 3\}$) to maintain computational efficiency and prevent the agent from accumulating contradictory hypotheses.

In each iteration $i$, the agent assumes three functional roles sequentially. First, based on the canonicalized inputs $(\hat{Q}, \hat{V})$, the retrieved Knowledge Notes, and any observations from previous iterations, it generates a candidate physical explanation $H_i$ together with a preliminary candidate answer $A_{\text{cand}}$, accompanied by a set of actionable queries $\mathit{Queries}_i$ encoded as \texttt{<info></info>} and \texttt{<attention></attention>} signals. Second, these queries are executed against the canonicalized visual set $\hat{V}$ to extract micro-facts $\mathit{Facts}_i$. To ensure objective grounding, each fact must strictly adhere to the \textit{Triadic Observation Template} defined in Section~\ref{sec:perception}. Third, a specialized validator module assesses whether $\mathit{Facts}_i$ provides sufficient evidence to support $A_{\text{cand}}$. If critical information is missing, the validator provides explicit negative feedback, marking specific pieces of evidence as ``unobtainable'' to refine the hypothesis in iteration $i+1$.

In scenarios where the information gap remains unclosed after the iteration limit, the system enters a \textit{Degenerative Inference Mode}. In this state, the agent supplements grounded evidence with its internal world knowledge (i.e., heuristic assumptions) to produce a final answer $A$. While this fallback ensures that every question receives an answer, the resulting reasoning is less trustworthy because it relies on unverified priors.

\paragraph{Update Eligibility and Reflection.} After the InfoAgent produces an answer, the system determines whether the resulting reasoning trace should be used to update the Knowledge Notes $\mathcal{N}$. We define a reasoning trace $P$ as the full record of hypotheses, queries, and observations across all iterations. During the training phase, where ground-truth labels are available, $P$ is eligible for knowledge consolidation only when it represents a fully grounded success, meaning four conditions must hold simultaneously: (i)~the final answer matches the ground truth, (ii)~the Degenerative Inference Mode was not activated, (iii)~no heuristic assumption was used in place of grounded evidence, and (iv)~the validated micro-facts contain explicit physical operators (e.g., \textit{trajectory, collision, contact}) rather than purely linguistic descriptors. Denoting these four predicates as $c$, $f$, $a$, and $d$ respectively, eligibility is expressed as their conjunction:
\begin{equation}
\Phi(P) \;=\; c(P) \;\wedge\; \neg\, f(P) \;\wedge\; \neg\, a(P) \;\wedge\; d(P)
\end{equation}
This strict gate prevents the system from consolidating unverified biases into its long-term knowledge.
The reflection phase concludes with a dual-path update logic. If $P$ is both correct and decisive, the system performs a reflective update to the \textit{Task Details} and \textit{General Tips} within $\mathcal{N}$, extracting the underlying physical principles from the trace. Conversely, if the model fails but a ground truth exists, a \textit{Tip Discovery} routine is initiated to propose 1--2 actionable, observable tips that are appended to the corresponding \textit{Task Details} for future reuse. This reflective cycle ensures that the system learns not only from its grounded successes but also from its perceptual and logical failures.

\section{Experiments}

\subsection{Implementation Details}


For implementation, we employ Qwen2.5-VL-72B-Instruct~\cite{bai_2025_qwen25vl} as the base. Video inputs are processed by sampling 4 frames uniformly across the sequence at a resolution of 512 pixels. Knowledge management leverages a hierarchical \texttt{JSON} structure and \texttt{all-MiniLM-L6-v2} embeddings for task retrieval. To maintain knowledge purity, the pruning mechanism deprecates any node $T$ with an error rate $E(T) >0.7$ after at least 8 activations. The \textit{InfoAgent} reasoning is capped at 3 iterations, utilizing a 15-primitive keyword filter to gate the Update Eligibility criterion $\Phi(P)$.

\subsection{Dataset and Baseline}
\subsubsection{Dataset Setup}
We evaluate PhysNote on PhysBench \cite{chow2025physbench}, a large-scale interleaved video-image-text dataset. For the knowledge evolution phase, we sample 500 entries from the official training set. Main quantitative results are reported on the full test set (10,000 entries), while ablation studies are conducted on the validation set (200 entries). All subsets maintain an equal distribution across the four major physical domains: 
(S1) \textbf{Physics-based dynamics} (e.g., collisions and projectile motion), 
(S2) \textbf{Physical object relationships} (e.g., spatial layout and support relations), 
(S3) \textbf{Physical scene understanding} (e.g., lighting and environment context), and 
(S4) \textbf{Physical object properties} (e.g., mass, friction, and elasticity).




\subsubsection{Baselines}
We compare \textit{PhysNote} against a diverse set of baselines categorized into three groups: (1) \textbf{Open-source VLMs}, including LLaVA-series~\cite{liu_2023_improved}, InternVL1.5~\cite{chen_2024_how}, and Mantis~\cite{jiang_2024_mantis}; (2) \textbf{Closed-source VLMs}, such as GPT-4o~\cite{openai_2024_gpt4o}, Gemini-1.5~\cite{reid_2024_gemini}, and Claude-3.5~\cite{anthropic_2024_the}; and (3) \textbf{Multi-agent systems}, specifically PhysAgent~\cite{chow2025physbench}, to evaluate the effectiveness of our knowledge-evolution framework. To establish an upper bound for physical reasoning, we also include human performance as a reference. All models are evaluated under a zero-shot setting to ensure a fair comparison of their inherent physical understanding and reasoning capabilities.

\subsection{Quantitative Analysis}
\subsubsection{Main Results}
Table~\ref{tab:vlm_comparison} presents the performance of various models on the PhysBench dataset. Our proposed \textit{PhysNote} consistently outperforms all baseline models across all four physical domains (S1--S4). Specifically, \textit{PhysNote} achieves an average score of 56.68\%, which is a 4.96\% absolute improvement over the state-of-the-art multi-agent baseline, \textit{PhysAgent}.

Compared to individual open-source and closed-source VLMs, the advantage of our framework is even more significant. While closed-source models like GPT-4o and Gemini-1.5-Pro show competitive results in specific categories, they often struggle with temporal consistency in dynamic scenes (S1) and complex object relationships (S2). In contrast, \textit{PhysNote} shows a balanced performance gain, particularly in Physics-based dynamics (S1) and Physical object relationships (S2). These results suggest that externalizing physical knowledge into structured notes effectively helps the model bridge the gap between raw visual inputs and high-level physical reasoning.

\begin{table}[ht]
\centering
\caption{\textbf{Quantitative results on PhysBench.} Accuracy comparison across open-source, closed-source, and multi-agent systems on four physical reasoning categories (S1--S4). \textbf{Bold} and \underline{underline} indicate the best and second-best results within each category.}
\label{tab:vlm_comparison}
\vspace{2mm}
\begin{small}
\begin{tabularx}{\columnwidth}{l l YYYYY}
\toprule
\rowcolor{header_color}
\textbf{Category} & \textbf{Model} & \textbf{S1} & \textbf{S2} & \textbf{S3} & \textbf{S4} & \textbf{Avg.} \\
\midrule
\textbf{Baselines} & Human & 97.10 & 95.67 & 94.91 & 95.68 & 95.87 \\
\midrule
\multirow{6}{*}{\textbf{Open-source}} & LLaVA-1.5-13B & 41.31 & 42.50 & 34.40 & 44.38 & 40.45 \\
& LLaVA-1.6-vicuna-7B & 40.26 & \underline{59.72} & \textbf{38.60} & 42.65 & 42.28 \\
& InternVL-Chat-1.5 & \textbf{53.08} & \textbf{70.14} & 37.01 & \underline{44.78} & \textbf{47.51} \\
& PLLaVA-13B & 39.91 & 38.33 & 31.52 & 40.76 & 37.70 \\
& Mantis-siglip-llama3 & 42.47 & 32.78 & \underline{36.83} & 37.51 & 37.64 \\
& LLaVA-interleave-dpo & 47.97 & 42.67 & 33.73 & 38.78 & 40.83 \\
\midrule
\multirow{4}{*}{\textbf{Closed-source}} & GPT-4o & 56.91 & \textbf{64.80} & 30.15 & \textbf{46.99} & \textbf{49.49} \\
& Gemini-1.5-pro & \underline{57.26} & \underline{63.61} & \textbf{36.52} & 41.56 & \underline{49.11} \\
& Claude-3.5-sonnet & 46.46 & 41.11 & 27.89 & 37.60 & 38.05 \\
& Gemini-1.5-flash & \textbf{57.41} & 52.24 & 34.32 & 40.93 & 46.07 \\
\midrule
\multirow{2}{*}{\textbf{Multi-agent}} & PhysAgent & 58.20 & 65.40 & \underline{38.10} & \underline{45.20} & 51.72 \\
& \cellcolor{best_cell} \textbf{PhysNote (Ours)} & \cellcolor{best_cell} \textbf{62.45} & \cellcolor{best_cell} \textbf{72.10} & \cellcolor{best_cell} \textbf{42.33} & \cellcolor{best_cell} \textbf{49.85} & \cellcolor{best_cell} \textbf{56.68} \\
\bottomrule
\end{tabularx}
\end{small}
\end{table}

\subsubsection{Ablation Results}
\begin{table}[ht]
\centering
\caption{\textbf{Ablation study on PhysBench validation set.} Using Qwen2.5-VL-72B-Instruct as the base model, we evaluate the contribution of each component across four task categories. \textbf{Bold} indicates the best result; \underline{underline} indicates the second best.}
\label{tab:ablation_results}
\vspace{2mm}
\begin{small}
\begin{tabular}{l | cccc | c}
\toprule
\rowcolor{header_color}
\textbf{Model Configuration} & \textbf{Dynamics} & \textbf{Scene} & \textbf{Relationships} & \textbf{Property} & \textbf{Overall Avg.} \\
\midrule
Baseline (VLM only) & \underline{70.89} & 54.05 & \underline{73.91} & \textbf{78.38} & \underline{69.85} \\
Baseline+InfoAgent  & 63.29  & 54.05 & 71.74  & 67.57 & 64.32   \\
Baseline+Note & 60.76  & \underline{56.76} & \textbf{78.26}  & \underline{75.68}  & 66.83 \\
\midrule
\cellcolor{best_cell}\textbf{Ours (Full)} & \cellcolor{best_cell}\textbf{74.68} & \cellcolor{best_cell}\textbf{62.16} & \cellcolor{best_cell}\textbf{78.26} & \cellcolor{best_cell}72.97 & \cellcolor{best_cell}\textbf{72.86} \\
\bottomrule
\end{tabular}
\end{small}
\end{table}

Table~\ref{tab:ablation_results} shows the contribution of each component to the overall performance. We observe that simply adding the \textit{InfoAgent} or \textit{Knowledge Notes} alone does not always lead to immediate improvements. In fact, the ``Baseline+InfoAgent'' configuration shows a slight performance drop, likely because the agent acts as a stochastic parrot when it lacks structured guidance, relying on semantic correlations rather than true physical grounding.

However, when these components are integrated into our Full Model, we achieve the best performance in most categories, with an overall accuracy of 72.86\%. This demonstrates that the synergy between agentic exploration and structured knowledge management is essential for solving complex physical tasks.
\begin{figure}[H]
    \centering
    \small
    \begin{tabular}{@{}cccc@{}}
        \includegraphics[width=0.24\textwidth]{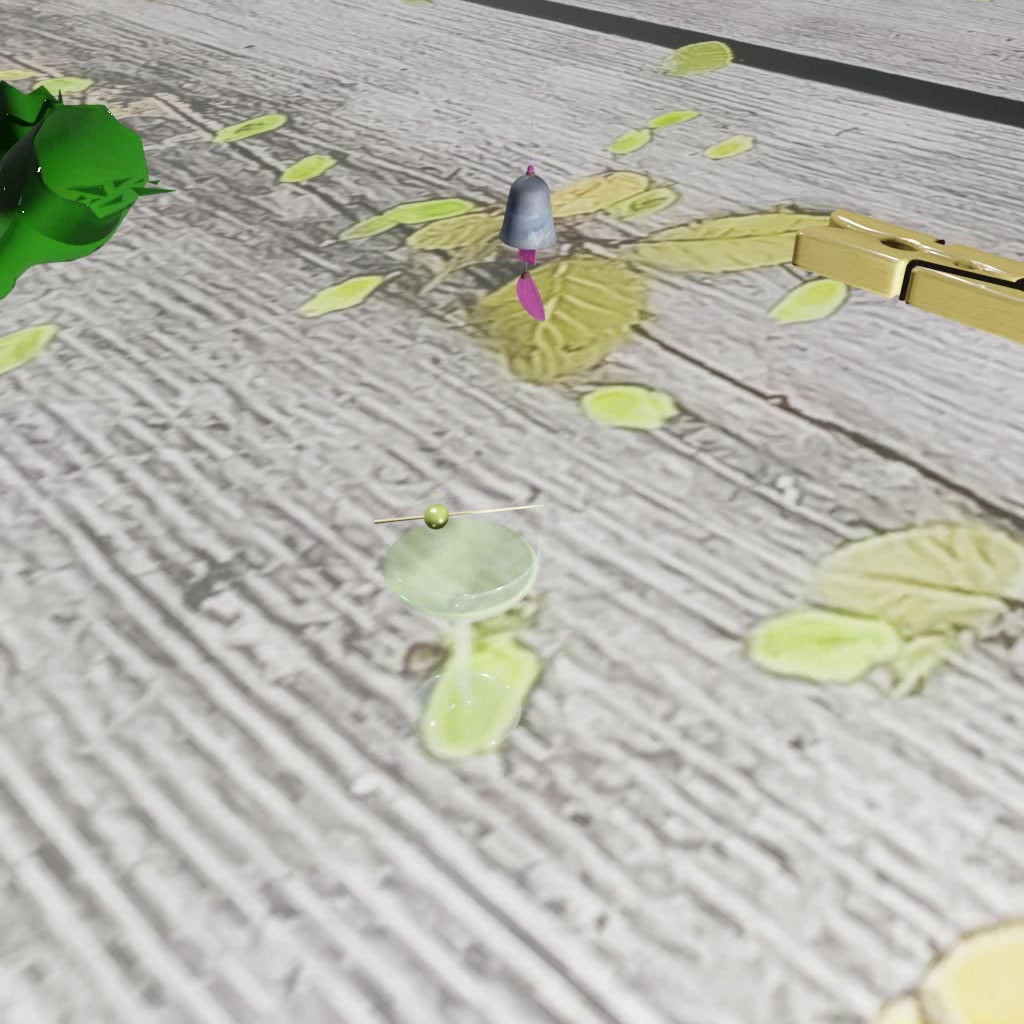} & 
        \includegraphics[width=0.24\textwidth]{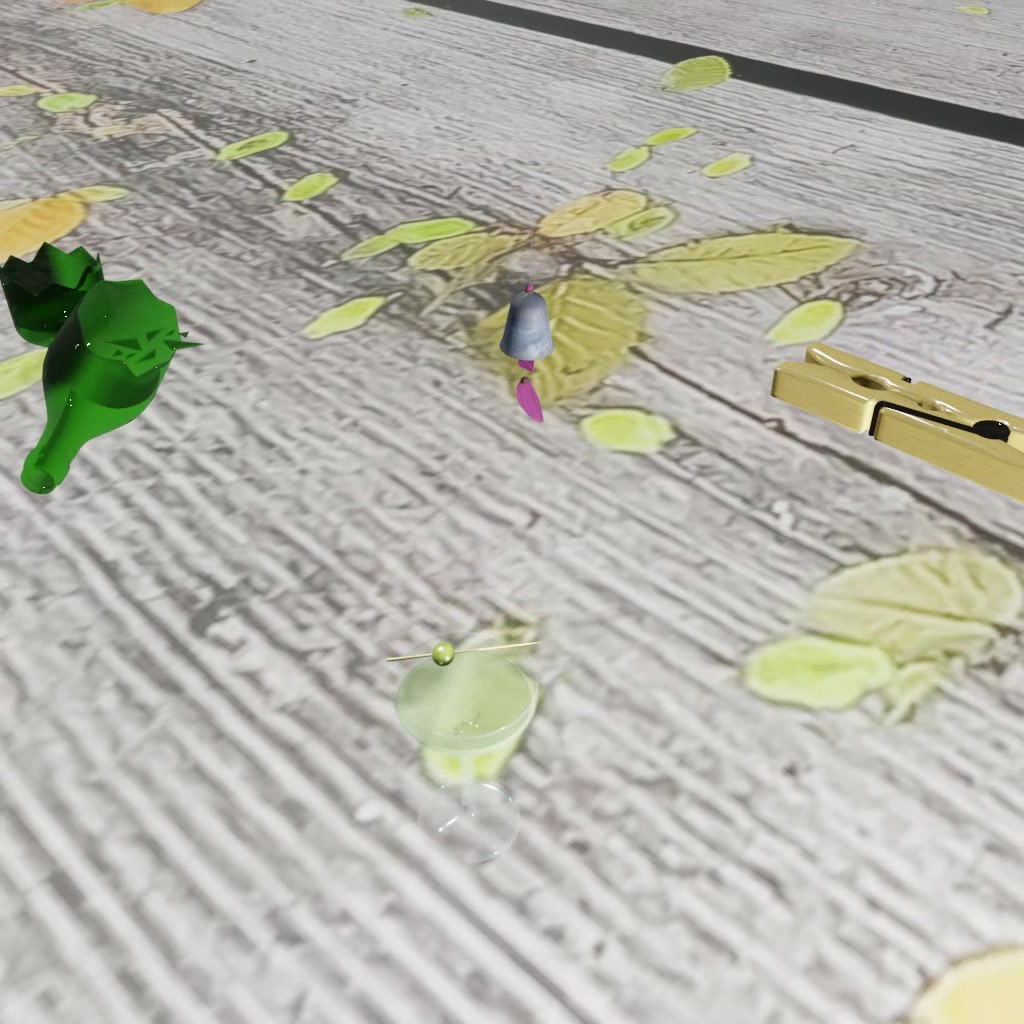} & 
        \includegraphics[width=0.24\textwidth]{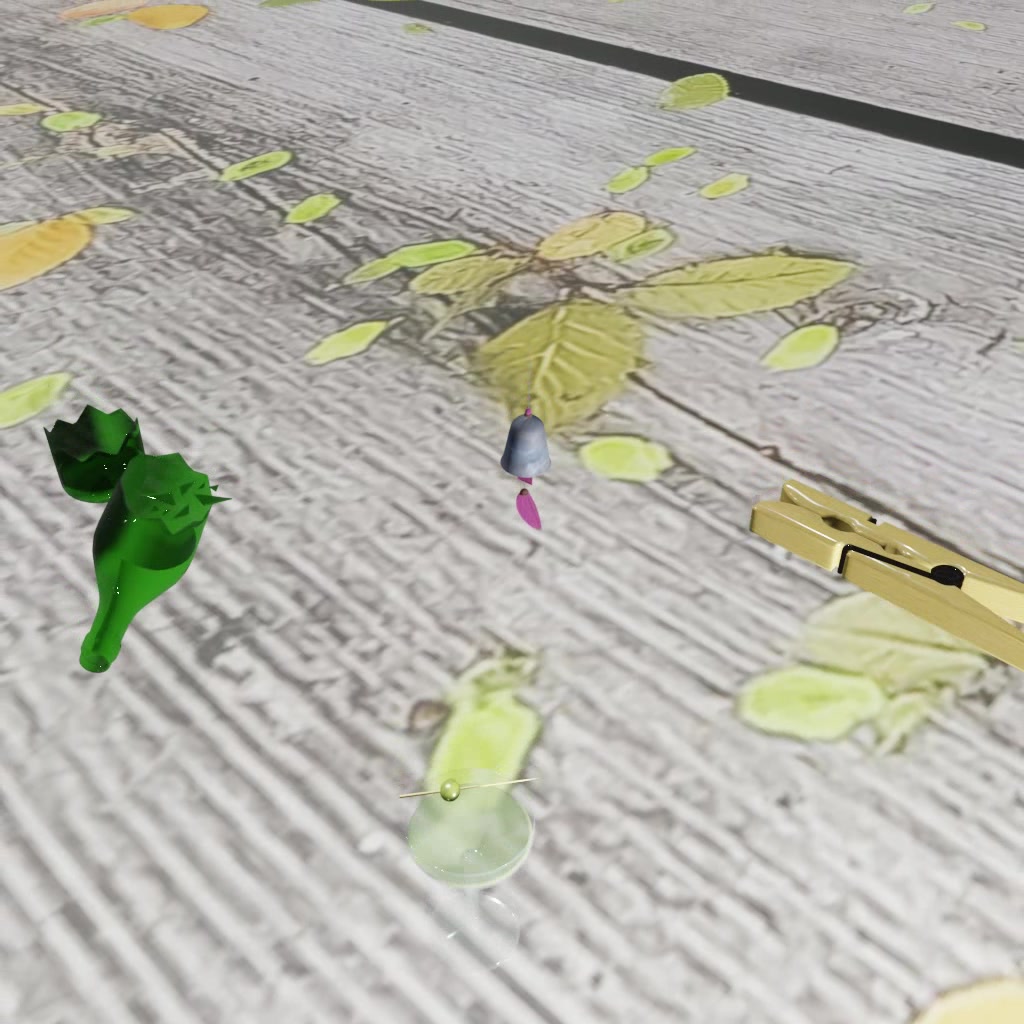} & 
        \includegraphics[width=0.24\textwidth]{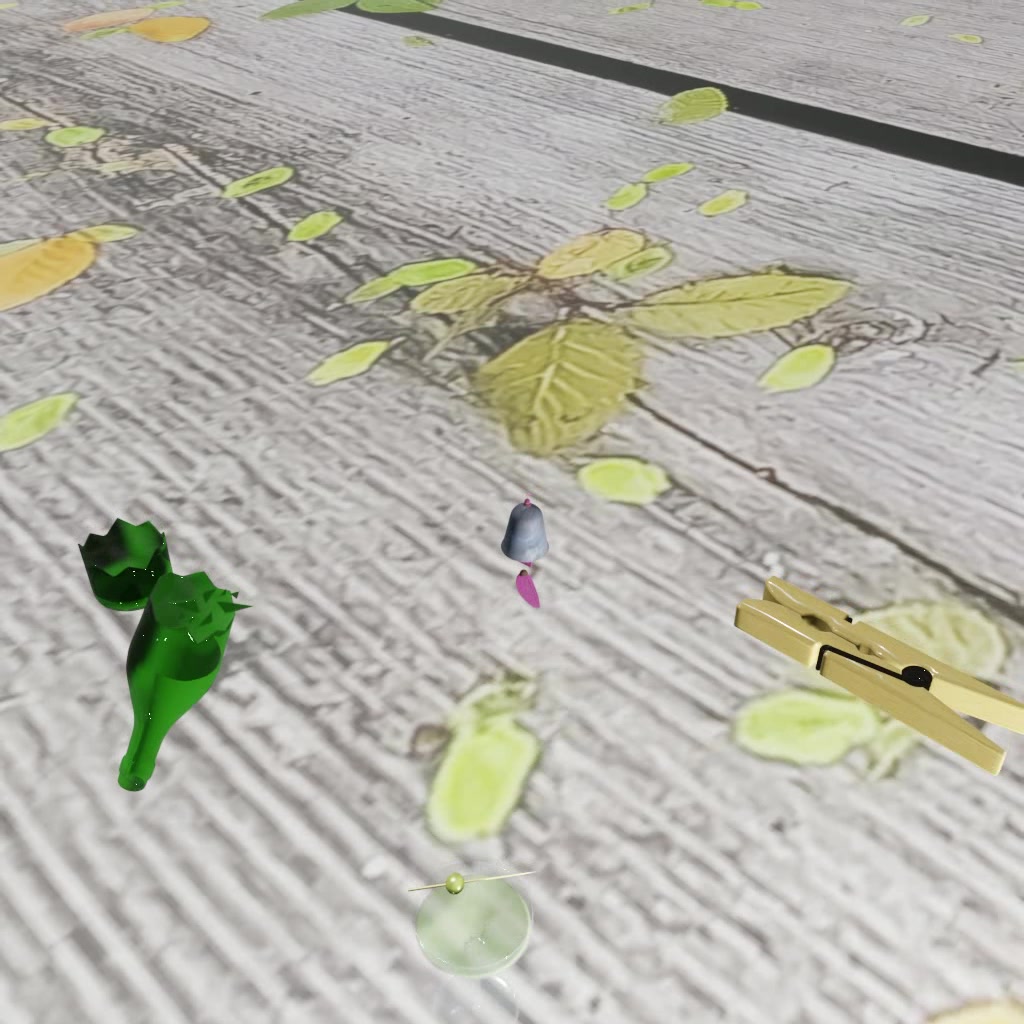} \\
        \addlinespace[5pt]
        \multicolumn{4}{p{0.96\textwidth}}{
            \textbf{Question:} Given the observed phenomena in the video, which of the following camera adjustments could create the depicted effect? 
            A. Move parallel to the line between the broken wine bottle and the martini; 
            B. The camera is farther away from the objects; 
            C. Move parallel to the line between the peg and the martini; 
            D. The camera moves upward or downward. \par
            \textbf{Ground Truth: D}
        } \\ \midrule
    \end{tabular}

    \vspace{1mm}

    \fontsize{8pt}{10pt}\selectfont
    \begin{tabular}{@{}p{0.5\textwidth}|p{0.52\textwidth}@{}}
        \centering \small\textbf{Baseline} & \centering \small\textbf{Ours} \tabularnewline \hline
        \textbf{Raw Reasoning:} 
        Analyze changes in video frames: 
        1. \textbf{Objects' Positions}: Relative positions appear consistent. 
        2. \textbf{Perspective Changes}: No significant change in angle, suggests camera is not moving closer/farther. 
        3. \textbf{Vertical Alignment}: Objects maintain alignment, indicating camera is not moving upward/downward. 
        
        \textit{Observations:} 
        - Opt A: Parallel move wouldn't significantly alter scene. 
        - Opt B: Farther away would show size change (not observed). 
        - Opt C: Parallel move between peg and martini seems plausible. 
        - Opt D: Moving upward/downward would change vertical alignment (not observed). & 
        \textbf{Raw Reasoning:} 
        Tips applied: Consistent relative size observed between bottle/martini; stable camera distance. Peg/martini alignment stable; no parallel movement.
        
        \textbf{Iteration 1-2:} Hypothesis: bottle/martini maintain size; peg/martini show no change in alignment. Gathered Truths: Peg's tip aligns consistently with top rim. Fluid surface below rim appears constant.
        
        \textbf{Iteration 3:} Comparison of relative sizes shows consistency (rules out Opt B). Peg/martini alignment constant (rules out Opt C). Gray bell-shaped object distance to edge decreased. 
        
        \textbf{Assumption Rationale:} Consistent alignment and stable sizes suggest unchanged distance. Gray object position supports the inference for vertical displacement. \tabularnewline
        \hline
        \textbf{Extract Answer:} \textcolor{red}{\textbf{C}} & \textbf{Extract Answer:} \textcolor{blue}{\textbf{D}} \tabularnewline
    \end{tabular}
    
    \caption{\textbf{Qualitative comparison on camera motion reasoning.} Given a video sequence and a multiple-choice question about camera adjustments, the baseline (left) fails to detect the vertical shift and relies on perceived stability to select a lateral motion (C). Our method (right) iteratively verifies geometric constraints, including object alignment and relative scale, across successive frames. By ruling out lateral and depth-based hypotheses through grounded evidence, it correctly identifies vertical camera displacement (D).}
    \label{fig:raw_comparison}
\end{figure}

\subsection{Qualitative Analysis}

To further demonstrate the reasoning process, we present two representative cases comparing \textit{PhysNote} with the baseline.
\subsubsection{Case I: Evidence-Based Visual Anchoring }
 
 As illustrated in Figure~\ref{fig:raw_comparison}, the task requires identifying vertical camera displacement within a sequence of video frames. The Baseline model demonstrates a failure to maintain precise spatial tracking across the temporal sequence, leading to a misperception of stable vertical alignment. This perceptual drift causes the model to overlook subtle geometric shifts and instead rely on a plausible but physically incorrect lateral motion (Option C). In contrast, PhysNote utilizes its iterative reasoning loop to systematically evaluate specific geometric constraints. By generating and verifying "Gathered Truths" regarding the relative size of objects and the alignment of the peg’s tip, the framework rules out hypotheses involving depth and parallel movement. Specifically, by detecting the displacement of the gray object relative to the frame boundary, PhysNote correctly identifies the vertical motion (Option D). This case demonstrates that by anchoring perception through structured knowledge notes, the model can transition from passive pattern matching to an evidence-based verification process, thereby mitigating the identity drift prevalent in standard vision-language reasoning

\subsubsection{Case II: Knowledge-Guided Attribute Discrimination}
\begin{figure}[H]
    \centering
    \small
    \begin{tabular}{@{}cccc@{}}
        \includegraphics[width=0.24\textwidth]{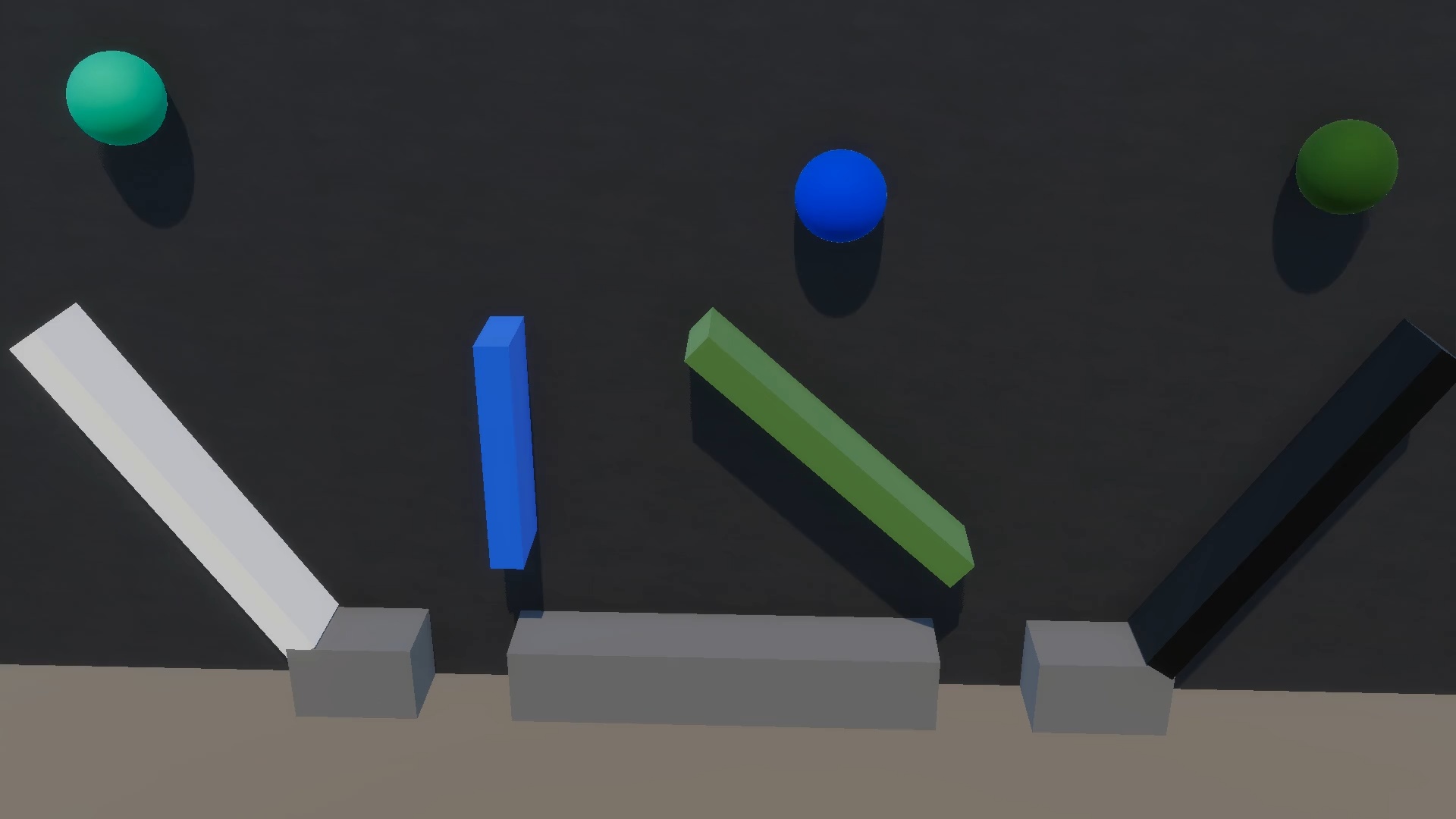} & 
        \includegraphics[width=0.24\textwidth]{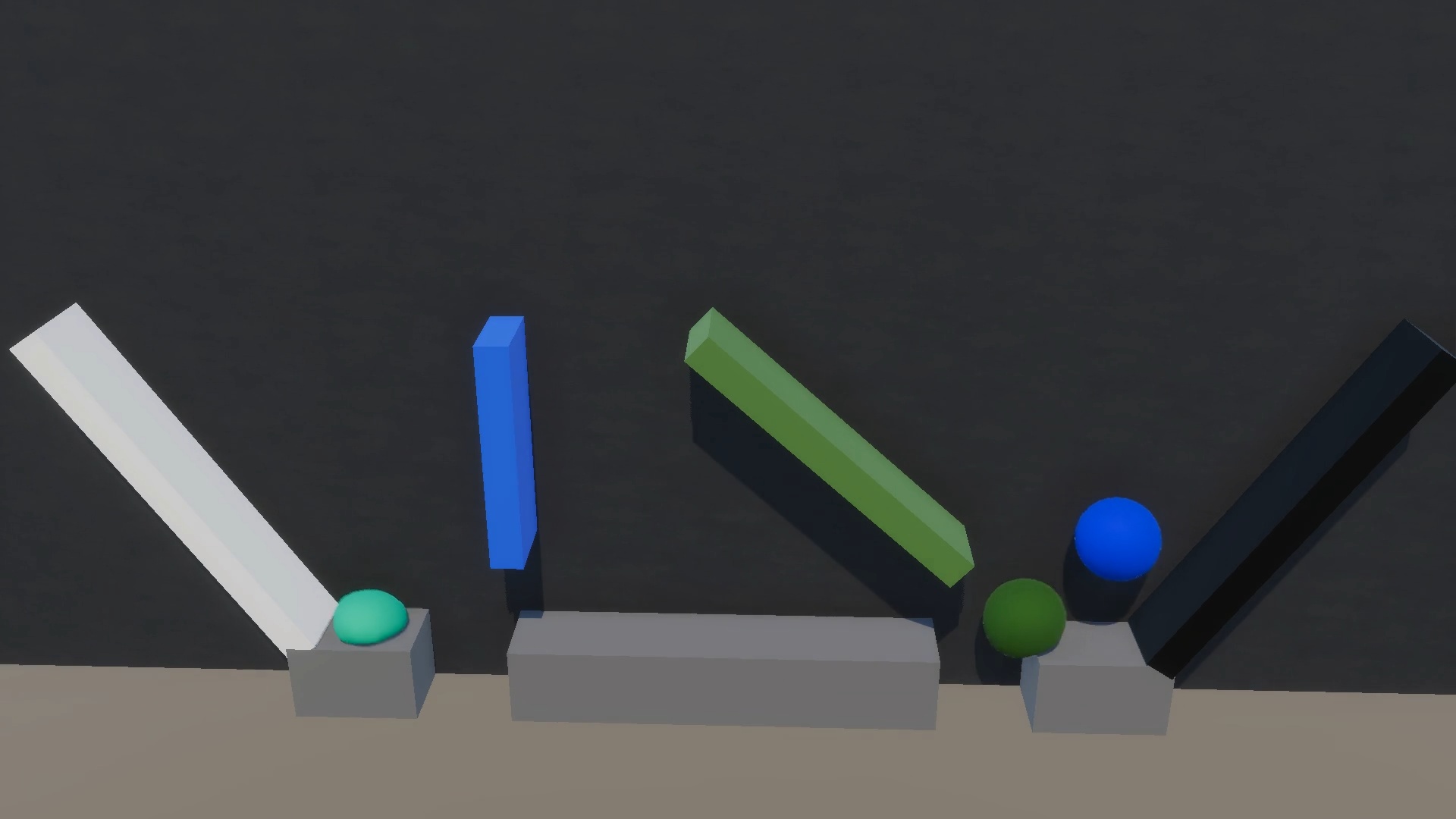} & 
        \includegraphics[width=0.24\textwidth]{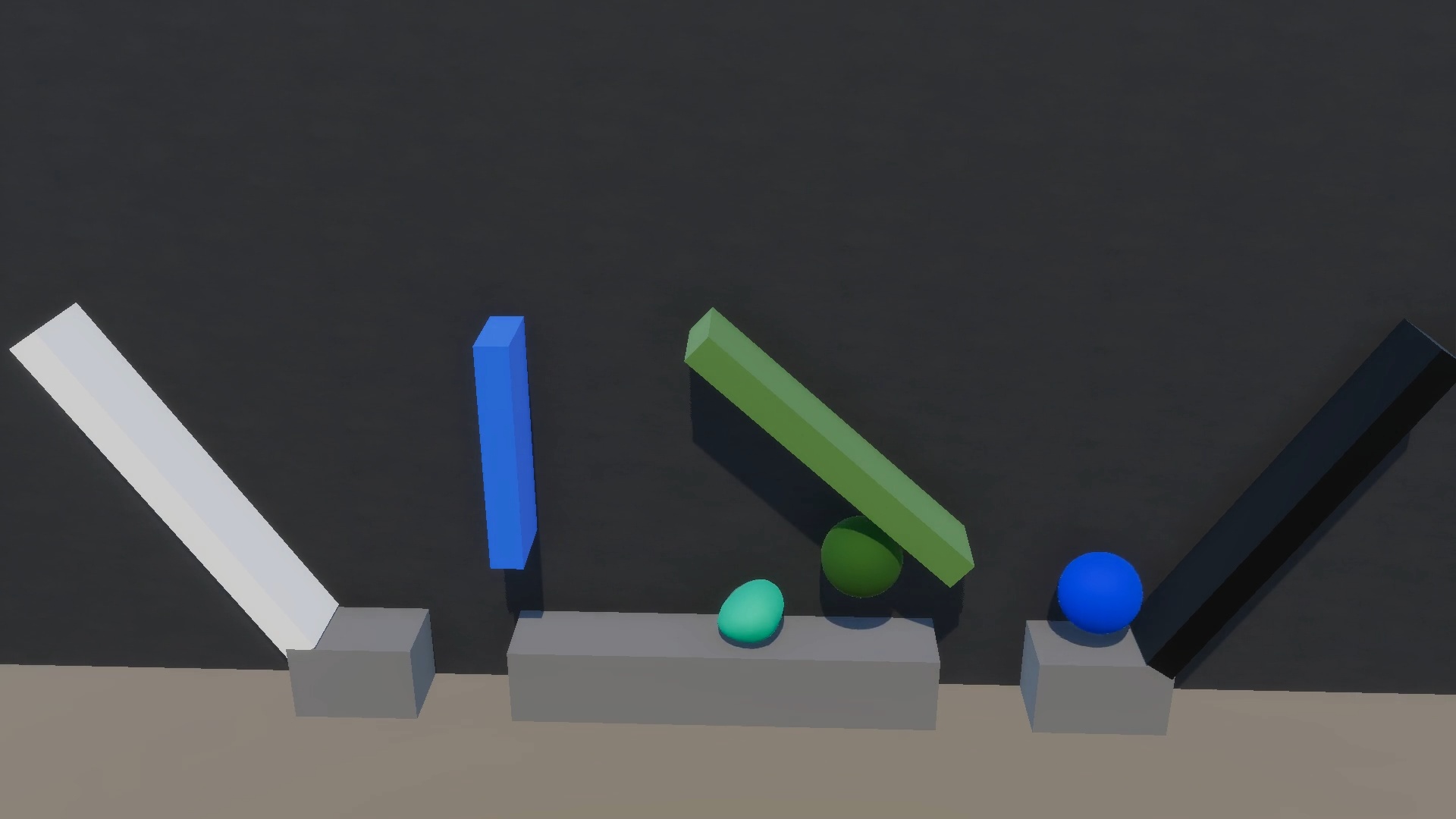} & 
        \includegraphics[width=0.24\textwidth]{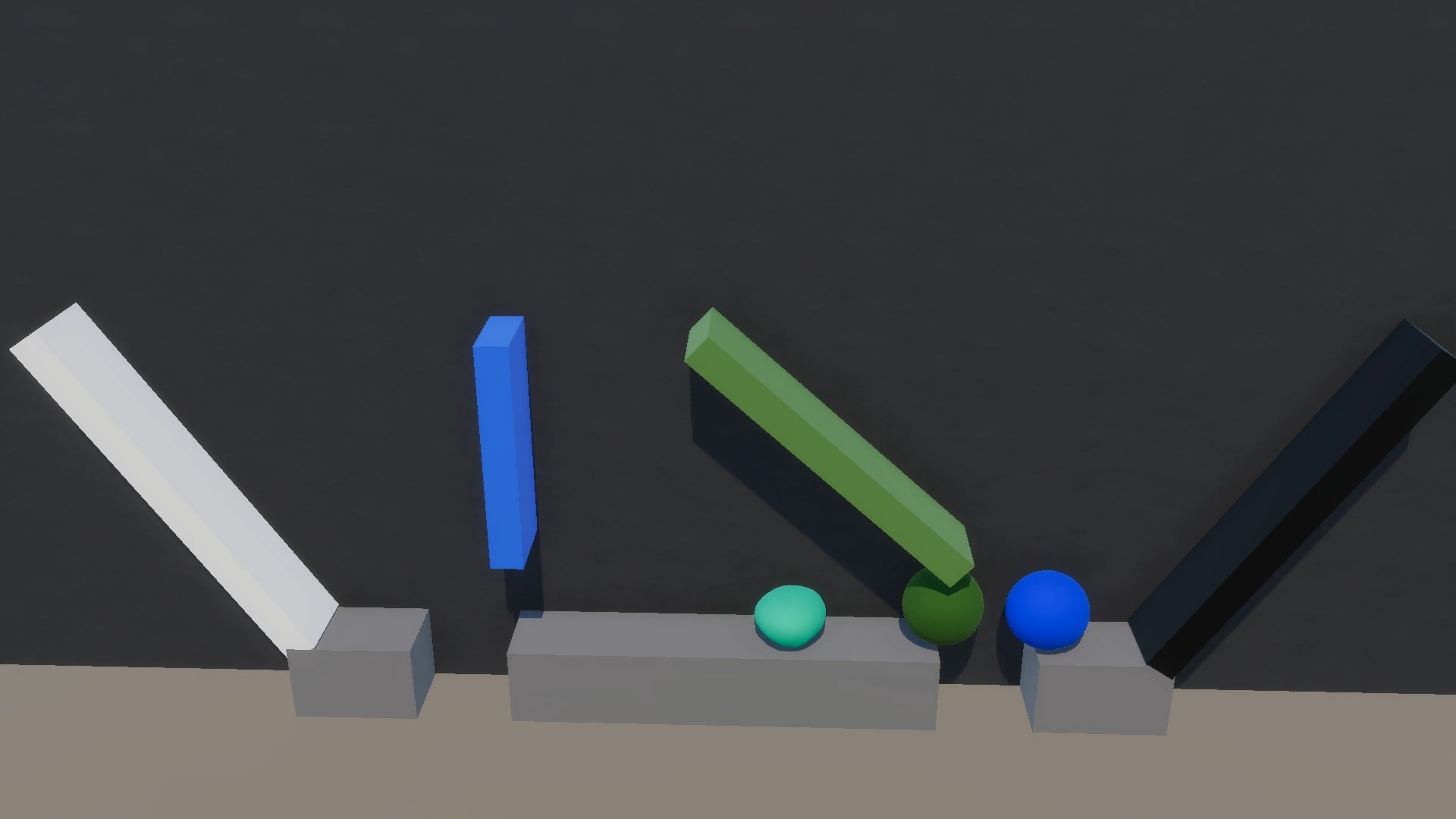} \\
        \multicolumn{4}{p{0.96\textwidth}}{\textbf{Question:} How does the plasticity of the green ball compare to the cyan ball? A. Much less B. Much more C. About the same D. Uncertain \par
            \textbf{Ground Truth: A}}  \\
        \midrule
        
    \end{tabular}

    \vspace{1mm}

    \fontsize{8pt}{10pt}\selectfont
    \begin{tabular}{@{}p{0.5\textwidth}|p{0.52\textwidth}@{}}
        \centering \small\textbf{Baseline} & \centering \small\textbf{Ours} \tabularnewline \hline
        \textbf{Raw Reasoning:} 
        To determine the plasticity, let's observe their behavior:
        1. \textbf{Initial Position}: Both balls are at the top of their respective ramps.
        2. \textbf{Movement}: As the balls roll down, they appear to maintain their spherical shape without deforming significantly.
        3. \textbf{Final Position}: Both balls reached the bottom and are resting. Neither ball shows any noticeable deformation or change in shape.
        
        \textit{Conclusion:}
        - Both balls retain spherical shape throughout.
        - No visible difference in how they deform as they move down and come to rest.
        Therefore, plasticity is about the same. & 
        \textbf{Raw Reasoning:} 
        Tips applied: If the shadow of a ball is sharper and smaller (e.g., green ball), consider lower plasticity, because less material displacement indicates less deformation. If a ball maintains a more spherical shape after interaction, consider lower plasticity.
        
        \textbf{Iteration 1:} 
        - \textit{Hypothesis:} Check if green sphere's shadow is sharper/smaller than cyan's in frame 0. Check if green retains a more spherical shape in frame 3 after surface interaction.
        - \textit{Gathered Truths:} Green sphere shadow is sharper/smaller. Green sphere retains a more spherical shape (approx. 15\% less deformation than cyan).
        
        \textbf{Validation:} Sharp shadow and spherical retention support lower plasticity for the green sphere. \tabularnewline
        \hline
        \textbf{Extract Answer:} \textcolor{red}{\textbf{C}} & \textbf{Extract Answer:} \textcolor{blue}{\textbf{A}} \tabularnewline
    \end{tabular}
    
    \caption{\textbf{Qualitative comparison on physical property reasoning.} The baseline (left) observes no obvious deformation in either ball and concludes that their plasticity is equal (C). Our method (right), guided by retrieved Knowledge Notes on material cues, leverages shadow sharpness and post-interaction shape retention as observable proxies for plasticity. By quantifying the difference in deformation between the two balls, it correctly identifies that the green ball exhibits much less plasticity (A).}
    \label{fig:plasticity_case}
\end{figure}

Figure~\ref{fig:plasticity_case} illustrates a scenario requiring the comparison of plasticity between two rolling balls. In this case, the \textbf{Baseline} failure shifts from spatial tracking to causal depth. While the baseline performs a full-motion analysis, it concludes that both balls are identical due to their similar macro-motion. This failure stems from ``not knowing what to look for''---the model lacks the specific physical priors to prioritize subtle material-related signals.
Conversely, \textit{PhysNote} uses its \textbf{Note system} to retrieve task-specific ``Tips'' from the knowledge base. These notes direct the agent to focus on fine-grained indicators, such as the sharpness of shadows and the degree of spherical retention (identifying a 15\% difference in deformation). By externalizing these physical insights, \textit{PhysNote} successfully differentiates the materials. This case shows that our framework allows the model to analyze physically meaningful signals that are typically ignored by baseline.

\section{Conclusion}
We present \textit{PhysNote}, a framework that enhances VLM physical reasoning through externalized Knowledge Notes. By integrating spatio-temporal canonicalization with a hierarchical knowledge structure, we effectively reduce identity drift and reasoning volatility in dynamic scenes. Experimental results on PhysBench show that \textit{PhysNote} achieves 56.68\% and 72.86\% accuracy on the test and validation sets, respectively. 

\clearpage
\bibliography{iclr2026_conference}
\bibliographystyle{iclr2026_conference}


\end{document}